\documentclass{article}

\usepackage{arxiv}

\usepackage[utf8]{inputenc} % allow utf-8 input
\usepackage[T1]{fontenc}    % use 8-bit T1 fonts
\usepackage{url}            % simple URL typesetting
\usepackage{booktabs}       % professional-quality tables
\usepackage{amsfonts}       % blackboard math symbols
\usepackage{nicefrac}       % compact symbols for 1/2, etc.
\usepackage{microtype}      % microtypography
\usepackage{lipsum}		% Can be removed after putting your text content
\usepackage{graphicx}
\usepackage{natbib}
\usepackage{doi}
\usepackage{tabularray}

\title{LDEB: Label Digitization with Emotion Binarization and Machine Learning for Emotion Recognition in Conversational Dialogues}

\author{ \href{https://orcid.org/0009-0006-3435-4730}{\includegraphics[scale=0.06]{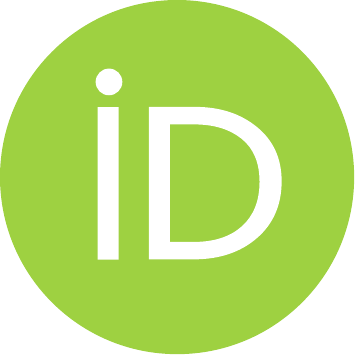}\hspace{1mm}Amitabha Dey}\\
	Department of Computer Science\\
	University of North Carolina at Greensboro\\
	North Carolina, NC 27412 \\
	\texttt{a\_dey@uncg.edu} \\
	\And
	\href{https://orcid.org/0000-0003-3235-9870}{\includegraphics[scale=0.06]{orcid.pdf}\hspace{1mm}Shan Suthaharan}\\
	Department of Computer Science\\
	University of North Carolina at Greensboro\\
	North Carolina, NC 27412\\
	\texttt{s\_suthah@uncg.edu} \\
}

\renewcommand{\shorttitle}{LDEB and ML for Emotion Recognition in Conversational Dialogues}

\hypersetup{
pdftitle={LDEB and ML for Emotion Recognition in Conversational Dialogues},
pdfsubject={q-bio.NC, q-bio.QM},
pdfauthor={Amitabha Dey, Shan Suthaharan},
}

\begin{document}
\maketitle

\begin{abstract}
Emotion recognition in conversations (ERC) is vital to the advancements of conversational AI and its applications. Therefore, the development of an automated ERC model using the concepts of machine learning (ML) would be beneficial. However, the conversational dialogues present a unique problem where each dialogue depicts nested emotions that entangle the association between the emotional feature descriptors and emotion type (or label). This entanglement that can be multiplied with the presence of data paucity is an obstacle for a ML model. To overcome this problem, we proposed a novel approach--called Label Digitization with Emotion Binarization (LDEB)--that disentangles the twists by utilizing the text normalization and 7-bit digital encoding techniques and constructs a meaningful feature space for a ML model to be trained. We also utilized the publicly available dataset--called the FETA-DailyDialog dataset--for feature learning and developed a hierarchical ERC model using random forest (RF) and artificial neural network (ANN) classifiers. Simulations showed that the ANN-based ERC model was able to predict emotion with the best accuracy and precision scores of about 74\% and 76\%, respectively. Simulations also showed that the ANN-model could reach a training accuracy score of about 98\% with 60 epochs. On the other hand, the RF-based ERC model was able to predict emotions with the best accuracy and precision scores of about 78\% and 75\%, respectively.
\end{abstract}

% keywords can be removed

\section{Introduction}
The development of an automated system for emotion recognition in conversations (ERC) is beneficial to many conversational AI applications, \citep{hazarika2021conversational, bhat2021adcofe}. The recent language model ChatGPT in the domain of conversational AI has shown the usefulness of an automated system for ERC, \citep{shahriar2023let,zhang2023complete}. Such a system can help advance research in many disciplines that include computational linguistics, neuroscience, and psychology, \citep{canales2014emotion,strapparava2008learning}. There has been a significant effort to understand the emotions in conversations and develop efficient computational techniques and machine learning classifiers for ERC using the information in conversational dialogues, \citep{huang2018automatic, huang2019ana}. For example, \citep{huang2018automatic}--assuming that the textual information in a dialogue does not deliver sufficient information--proposed an approach to supply emotion information \textit{a priori} at training. Subsequently,  \citep{huang2019ana} have also utilized the Long Short Term Memory networks (LSTM) architecture hierarchically--as an iterative model--to capture contextual emotional features so that the model can predict the emotions in textual dialogues.

Machine learning (ML) is a technique that can help us develop such an automated system to recognize emotions in a conversational dialogue by performing the classification of emotions. For example, \citep{binali2010computational} have adapted emotion theories, based on Ekman’s model and the OCC (Ortony/Clore/Collins) model, and developed a support vector machine (SVM) classifier for emotion recognition in a web blog data. A brief discussion on the Ekman model and the OCC model can be found in \citep{zad2021emotion}. Similarly, deep learning techniques have also been studied to detect emotions in textural dialogues by extracting sentiment and semantic feature descriptors \citep{chatterjee2019understanding}. They also utilized an LSTM model to extract sentiment and semantic feature descriptors and build their deep-learning models. As reported in \citep{acheampong2020text}, we also observed that the convolutional neural network (CNN) and recurrent neural network (RNN) have also been used in this research domain for developing ERC models.

While these approaches provide solutions to emotion detection from the text, we observed that they did not address the \textit{emotion entanglement} issues that were inherent in many conversational dialogues. Here, it is important to differentiate the usage of the term \textit{entaglement} in the NLP domain and understand the context in which it is used in our research. In \citep{huang2021disentangling}, the authors used the term to describe the interlaced relationship between semantic and syntactic information. In \citep{webson2020undocumented}, the authors used the term to describe how pre-trained models encode denotation and connotations as one intertwined representation. Furthermore, we see the usage of the term \textit{word entanglement} in \citep{matthews2018evaluating}, by which the authors described the relation between content and style of sentences when considering the transferability of a sequence of tokens. However, in our research, we use the term to explain how two dialogues may be made up of the same set of emotions, but in different order of appearance and/or frequency. 

A conversational dialogue is generally composed of several utterances, or sub-dialogues that express different emotions; hence, the true message of the dialogue could be incorrectly interpreted by the listener. The individual utterance, or sub-dialogue annotation in isolation cannot be utilized accurately unless it is considered in relation to its preceding and following sub-dialogues, collectively. Because the same sub-dialogue in another context might be annotated with a different emotion label depending on the context. When a human language expert manually annotates the dialogues, they annotated them based on the contextual meaning. But this intra-relationship between the sub-dialogues could be lost if they were not captured together.

A conversational dialogue with such a state of confusion can cause problems for the successful development of conversational AI models from such data. We called this problem an \textit{emotion entanglement}, because of the nested nature of the emotions and the twisted association between the feature descriptors and the emotion types. For example, consider a dialogue $d_1$ that has six sub-dialogues $(u_0,u_1,u_2,u_3,u_4,u_5)$ that express 4 emotions (0,1,2,4) out of 7 emotions (0,1,2,3,4,5,6), and it is labeled as follows: (4, 2, 0, 1, 0, 1). Let's also consider another dialogue $d_2$ that has six sub-dialogues $(v_0,v_1,v_2,v_3,v_4,v_5)$ and labeled as (0, 2, 0, 0, 4, 1), then it also expresses the same four emotions (0,1,2,4). Therefore, in this case, we called that "the 4 emotions (0,1,2,4) are entangled" in the dialogues $d_1$ and $d_2$. This entanglement can get more complicated with respect to the increase in the number of emotions and sub-dialogues present in the dialogue. In other words, the emotion entanglement leads to the labeling of two dialogues--that express the same emotion set--with distinct values.

In our study, we have defined emotion entanglement as the measure of the significance of the order of emotions in a dialogue and its association with the feature descriptors via a computational model. Hence, the proposed approach alleviates this problem by digitizing the labels based on the binarization of emotions in an ordered sequence $(e_0,e_1,e_2,e_3,e_4,e_5,e_6)$, where the presence and absence of each emotion $e_i$ is marked at its position $i$, where $i=0...6$. For the above example, the dialogue $d_1$ is labeled as (1, 1, 1, 0, 1, 0, 0) = 116, and dialogue $d_2$ is also labeled as (1, 1, 1, 0, 1, 0, 0) = 116. Hence, the tokens in the dialogues $d_1$ and $d_2$ can be mathematically associated with the same label that is represented by a single integer between 0 and 127 in the proposed approach.

In addition, ML techniques require reliable labeled datasets to develop trustworthy ML models for an ERC system. Hence, our proposed work addressed these issues. Fortunately, a dialogue dataset, called DailyDialog, has been recently developed and distributed for the research use \citep{li2017dailydialog}. Using this dataset, the developers of this dataset evaluated some of the existing techniques by dividing them into three groups, namely embedding-based, feature-based, and neural network-based similarity-response retrieval approaches. Most importantly, this dataset is manually labeled; hence, it can be used to train machine learning models that can predict emotions in a conversational dialogue. Subsequently, the authors of \citep{albalak2022feta} utilized this dataset and developed a dataset--called FEw-sample TAsk transfer (FETA) data set--with the hope of efficiently training the large language models (LLMs) that are capable of performing self-supervised learning on a large unlabeled data, \citep{chen2020big}. However, it is still a question if this dataset has sufficient information to train an ML model under a data paucity problem and develop an ERC system.

\begin{table*}[!t]
\centering
\setlength{\tabcolsep}{5pt}
\renewcommand{\arraystretch}{1.2}
\begin{tabular}{|l|c|}
\hline
\multicolumn{1}{|c|}{\textbf{Dialogues\_Text}} & \textbf{Dialogues\_Emotion} \\ \hline
\begin{tabular}[c]{@{}l@{}}The kitchen stinks.\\ I’ll throw out the garbage.\end{tabular} & 2 0 \\ \hline
\begin{tabular}[c]{@{}l@{}}So Dick, how about getting some coffee for tonight?\\ Coffee? I don’t honestly like that kind of stuff.\\ Come on, you can at least try a little, besides your cigarette.\\ What’s wrong with that? Cigarette is the thing I go crazy for.\\ Not for me, Dick\end{tabular} & 4 2 0 1 0 \\ \hline
\begin{tabular}[c]{@{}l@{}}Would you mind waiting a while?\\ Well, how long will it be?\\ I’m not sure. But I’ll get a table ready as fast as I can.\\ OK. We’ll wait.\end{tabular} & 0 0 0 4 \\ \hline
\begin{tabular}[c]{@{}l@{}}What kind of food do you like?\\ I like Chinese food.\\But you're American?\\ We have a lot of Chinese restaurants in America.\end{tabular} & 0 4 6 0 \\ \hline
\begin{tabular}[c]{@{}l@{}}Is everything to your satisfaction?\\ No, the steak was recommended, but it is not very fresh.\\ Oh! Sorry to hear that. This is quite unusual as we have steak from the market every day.\\ So what? It is not fresh and I'm not happy about it.\\ I'm sorry, sir. Do you wish to try something else? That would be on the house, of course.\\ No, thank you.\end{tabular} & 0 1 5 1 5 1 \\ \hline
\begin{tabular}[c]{@{}l@{}}Do you like snakes?\\ Of course not. I'm afraid of them.\\ I mean, do you like snake meat?\\ I'm afraid I'll feel sick at the sight of it.\end{tabular} & 0 3 0 2 \\ \hline
\end{tabular}%
\caption{Every sub-dialogue of a dialogue is annotated with one of the following emotion classes - 0: no emotion, 1: anger, 2: disgust, 3: fear, 4: happiness, 5: sadness, and 6: surprise.
}
\label{table:multiclass-labels}
\end{table*}

In this paper--to address the problems and challenges caused by the emotion entanglement and data paucity--we proposed an approach to generate a feature space that consists of feature vectors with disentangled emotions and feature descriptors. We called this approach the "Label Digitization with Emotion Binarization (LDEB)" and utilized the text normalization and 7-bit digital encoding techniques to disentangle the twist of the association between the feature descriptors and emotion types. We also used the FETA-DailyDialog dataset to study our proposed approach. While the proposed LDEB approach offers a solution to address this problem, it can also introduce a new imbalanced data problem. Hence, it restricts the direct application of an ML technique, \citep{suthaharan2016machine}. To alleviate this problem, we proposed a hierarchical ML solution that will be discussed in the subsequent sections. Our proposed work also offered a refined dataset--we called the LDEB-DailyDialog dataset--that is a derivative of the DailyDialog \citep{li2017dailydialog} and FETA-DailyDialog datasets \citep{albalak2022feta}. The refined LDEB-DailyDialog dataset is a systematically organized feature space that enables its direct utilization for training conversational AI models for emotion recognition.

\section{Proposed Methodology}
\label{sec:proposedmethodology}
The proposed methodology consists of four modules. The first module integrates the proposed concept of Label Digitization with Emotion Binarization which generates a set of meaningful aggregated emotions for a dialogue. The second module performs a feature learning technique that generates feature vectors for conversational dialogues and maps the feature vectors to the aggregated emotions. The third module presents a hierarchical data modeling that splits a given dataset into balanced subsets (we called them Split-Sets) of data for the training of ML classifiers in a hierarchically ordered sequence.  The fourth module assumes RF and ANN classifiers for the models used in the hierarchical data structure and performs simulations to show the feasibility and performance efficiency of the proposed LDEB machine learning model.

\subsection{Proposed LDEB Concept}
The FETA-DailyDialog dataset consists of information that allows a mapping between \textit{Dialogues\_Text} and \textit{Dialogues\_Emotion} that are useful for our goal of developing ML models for emotion recognition in conversations. The information that is useful for our study includes the seven types of emotions--anger, disgust, fear, happiness, sadness, surprise, and no emotion--and the number of instances (or sub-dialogues) associated with these classes. The number of sub-dialogues is 1,022, 353, 74, 12,885, 1,150, 1,823, and 85,572, respectively. However, there are 13,118 dialogues that are formed by these sub-dialogues. This label distribution shows the imbalanced nature of the DailyDialog dataset that can create problems and challenges in developing ML models to classify emotions using this dataset.  

\begin{table*}[t!]
\centering
\setlength{\tabcolsep}{10pt}
\renewcommand{\arraystretch}{1.5}
\begin{tabular}{|ccccccc|}
\hline
\multicolumn{1}{|c|}{\begin{tabular}[c]{@{}c@{}}Emo 0\\ (No Emotion)\end{tabular}} & \multicolumn{1}{c|}{\begin{tabular}[c]{@{}c@{}}Emo 1\\ (Anger)\end{tabular}} & \multicolumn{1}{c|}{\begin{tabular}[c]{@{}c@{}}Emo 2\\ (Disgust)\end{tabular}} & \multicolumn{1}{c|}{\begin{tabular}[c]{@{}c@{}}Emo 3\\ (Fear)\end{tabular}} & \multicolumn{1}{c|}{\begin{tabular}[c]{@{}c@{}}Emo 4\\ (Happiness)\end{tabular}} & \multicolumn{1}{c|}{\begin{tabular}[c]{@{}c@{}}Emo 5\\ (Sadness)\end{tabular}} & \begin{tabular}[c]{@{}c@{}}Emo 6\\ (Surprise)\end{tabular} \\ \hline
\multicolumn{1}{|c|}{64} & \multicolumn{1}{c|}{32} & \multicolumn{1}{c|}{16} & \multicolumn{1}{c|}{8} & \multicolumn{1}{c|}{4} & \multicolumn{1}{c|}{2} & 1 \\ \hline
\multicolumn{1}{|c|}{1} & \multicolumn{1}{c|}{0} & \multicolumn{1}{c|}{0} & \multicolumn{1}{c|}{0} & \multicolumn{1}{c|}{1} & \multicolumn{1}{c|}{0} & 0 \\ \hline
\multicolumn{7}{|c|}{Emo\_Sum = $2^{6} + 2^{2} = 64 + 4 = 68$} \\ 
\hline
\end{tabular}%
\caption{For each dialogue, Emo\_Sum is calculated using the LDEB approach. A Emo\_Sum of 64 means that the dialogue has the following emotion classes present - \textit{No Emotion} and \textit{Happiness}. A lookup table (Table \ref{tab:my-classes}) is further constructed for the 128 Emo\_Sum values and their corresponding combination of emotion classes.}
\label{tab:my-table}
\end{table*}

\begin{table*}[t]
\centering
\setlength{\tabcolsep}{10pt}
\renewcommand{\arraystretch}{1.5}
\begin{tabular}{|c|c|c|c|}
\hline
\textbf{Emo\_Sum} & \textbf{Binary Numbers} & \textbf{Description} & \textbf{Count} \\ \hline
4 & 0000100 & Happiness & 437 \\ \hline
64 & 1000000 & No Emotion & 6247 \\ \hline
65 & 1000001 & No Emotion + Surprise & 685 \\ \hline
66 & 1000010 & No Emotion + Sadness & 391 \\ \hline
68 & 1000100 & No Emotion + Happiness & 3708 \\ \hline
69 & 1000101 & No Emotion + Happiness + Surprise & 494 \\ \hline
80 & 1010000 & No Emotion + Disgust & 131 \\ \hline
96 & 1100000 & No Emotion + Anger & 278 \\ \hline
\end{tabular}%
\caption{It illustrates the binary representations of Emo\_Sum values, description, and counts. Only the Emo\_Sum values in the 5 groups generated by the hierarchical data modeling (as presented in Figure \ref{fig:hmodel}) are presented in the table. There is a total of 128 possible Emo\_Sum representations as per Table \ref{tab:my-table}.}

\label{tab:my-classes}
\end{table*}

One of the goals of the proposed approach is to alleviate the imbalanced nature of the data in the training set. In the DailyDialog dataset, the \textit{Dialogues\_Text} column contained 13,118 dialogues, as stated earlier in this paper. In the \textit{Dialogues\_Emotion} column, each sub-dialogue of dialogue was annotated with an emotion class representing the dominant emotion exhibited by the sub-dialogue. There were a total of 7 emotion classes. The emotion labels were as follows: 0: no emotion, 1: anger, 2: disgust, 3: fear, 4: happiness, 5: sadness, and 6: surprise. Each sentence of the dialogue is annotated with one of the 7 aforementioned emotion classes. Therefore, a dialogue, which is composed of multiple sub-dialogues, is annotated by several emotion classes. This environment presented a multi-class challenge since each dialogue has multiple corresponding emotions. Table \ref{table:multiclass-labels} shows an example of how every sub-dialogue has been annotated with one of the following seven emotions (0: no emotion, 1:anger, 2: disgust, 3: fear, 4: happiness, 5: sadness, and 6: surprise) in the FETA-DailyDialog dataset. Therefore, we have multi-class emotion labels for every dialogue. Table \ref{tab:my-table} shows an example of generating a label by our LDEB approach. It shows Emo\_Sum 68 represents no emotion and happiness, where Emo\_Sum is the variable that we defined to represent the newly defined combinations of emotions. This table also illustrates how Emo\_Sum values for every dialogue are computed using our LDEB approach.

There are a total of 7 emotion types so we represented them in a 7-bit binary encoding system. Every sub-dialogue has been annotated with one of these 7 emotion classes. We added 1 (if the emotion class is present) and 0 (if the emotion class is not present) in a particular dialogue. We performed binary addition to calculate the Emo\_Sum value. This value captures the emotional relationship between the sub-dialogues and is the prediction value for our LDEB-DailyDialog dataset. Table \ref{tab:my-classes} lists some of the Emo\_Sum representations of the possible combinations of emotions with the counts of dialogues for each Emo\_Sum.

\begin{figure*}[t]
\centering
\includegraphics[width=16cm]{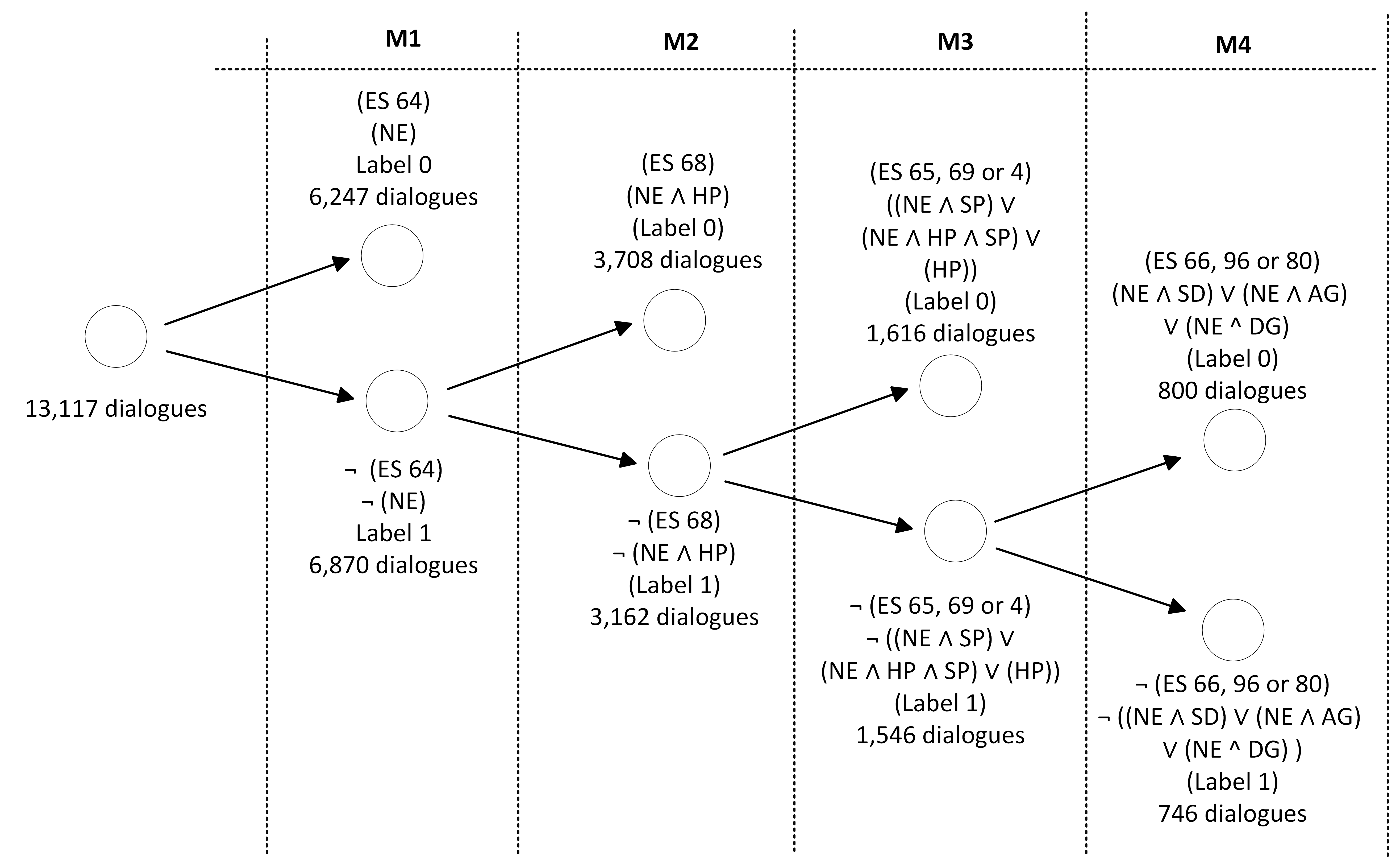}
\caption{The proposed hierarchical data modeling framework to improve the balanced nature of the data.}
\label{fig:hmodel}
\end{figure*}

\subsection{Feature Learning}
The proposed feature learning involved the extraction of emotional descriptors (features) and the association of features with Emo\_Sum values. It was performed by adapting the standard concept of text normalization that is used in natural language processing (NLP). We know that every dialogue is made up of sub-dialogues, and each sub-dialogue is composed of words. Hence, we utilized the bag-of-words (BoW) representation model to characterize these dialogues. We used each and every unique word as feature vectors and determined the occurrence frequency of each word in the respective dialogue. Initially, we tokenized all of the sentences in the 13,117 dialogues and then removed punctuation from the tokens. This process resulted in 1,200,389 words. We decided to keep the \textit{stop words} since the elimination of these words may result in the loss of important features that might be useful for machine learning. 

We also developed a word counter to address the word repetition problem. We created an empty list and added a new word to the list if it wasn't already present. The automation of this process identified 26,372 unique words in the 13,118 dialogues. Hence, our feature learning constructed a feature vector of dimension 26,372 where its feature is one of the unique words. By iterating through the dialogues, we examined the occurrence of these unique words in every dialogue and accordingly aggregated the tally to define the magnitude of the feature vectors for each dialogue, which is considered an observation of the feature space that we learned from the DailyDialog dataset.

\subsection{Hierarchical Data Modeling}
After constructing the feature space for \emph{LDEB-DailyDialog}, we found the dataset of 13,117 dialogues was still imbalanced as shown in Table \ref{tab:my-classes}. As we can see, there are 6,247 dialogues of Emo\_Sum value of 64 (Absence of Emotion) and 6,870 dialogues of other Emo\_Sum values (Presence of Emotion). Note that if a dialogue has "no emotion" class and "an emotion" (e.g., happiness or surprise) class, then the dialogue reflects an emotional dialogue (i.e., presence of emotion).

To address the imbalanced data problem, we adapted the concept of \emph{Hierarchical Machine Learning}. In this approach, we hierarchically built 4 machine-learning models by dividing (splitting) and grouping of dialogues with respect to their LDEB labels. Figure \ref{fig:hmodel} illustrates how the 4 subsets are split based on aggregating labels together to create more balanced sets. For model M1, we labeled all dialogues with an emotion sum (ES) value of 64 with Label 0. Dialogues with an ES value of 64 are dialogues that only have no emotion (NE) class annotated to all its sub-dialogues. There are 6,247 such dialogues. On the other hand, all dialogues that have ES values other than 64 are dialogues that have at least one sub-dialogue with an emotion class other than NE annotated ($\neg NE$). In other words, these dialogues are such that there is a presence of emotion. We labeled all of such dialogues as Label 1. There are 6,870 such dialogues as can be seen in  Figure \ref{fig:hmodel}. For building model M2, we discarded all the dialogues with an ES value of 64 that were previously considered for model M1. We labeled all dialogues with an ES value of 68 with Label 0. Dialogues with an ES value of 68 are dialogues that have both no emotion (NE) and happiness (HP) emotion classes annotated to their sub-dialogues. It is to be noted that these dialogues could have any number of sub-dialogues that are annotated with NE or HP. But by applying our LDEB approach, we are simply determining whether an emotion class is present in a dialogue or not. So the number of NE and HP annotations can be ignored. We have 3,708 dialogues with an ES value of 68. On the other hand, all dialogues that have ES values other than 68 are labeled as Label 1. There are 3,162 such dialogues.

For model M3, we discarded all the dialogues with an ES value of 68 that were previously considered in model M2. We labeled all dialogues with an ES value of 65, 69, or 4 with Label 0. These are collections of dialogues that have either one of the following characteristics: (a) Sub-dialogues only annotated with emotion class NE and surprise (SP) (b) Sub-dialogues only annotated with emotion classes NE and happiness (HP) and SP, or (c) Sub-dialogues only annotated with emotion class HP. There is no other motive for the selection of these emotion classes other than to create more balanced Split-Sets for model building. There are 1,616 such dialogues. On the other hand, all dialogues that have an ES value other than 65, 69, or 4 are labeled as Label 1. There are 1,546 such dialogues. Finally, for model M4, we discarded all the dialogues with an ES value of 65, 69, or 4 that were previously considered in model M3. We label all dialogues with an ES value of 66, 96, or 80 with Label 0. These are collections of dialogues that have either one of the following characteristics: (a) Sub-dialogues only annotated with NE and sadness (SD) (b) NE and anger (AG), or (c) NE and disgust (DG). There are 800 such dialogues. On the other hand, all dialogues that have an ES value other than 66, 96, or 80 are labeled as Label 1. There are 746 such dialogues. These interpretations of ES values can be further looked at in Table \ref{tab:my-classes}.

\begin{figure*}[ht!]
  \centering
  \includegraphics[width=16cm, height=6.5cm]{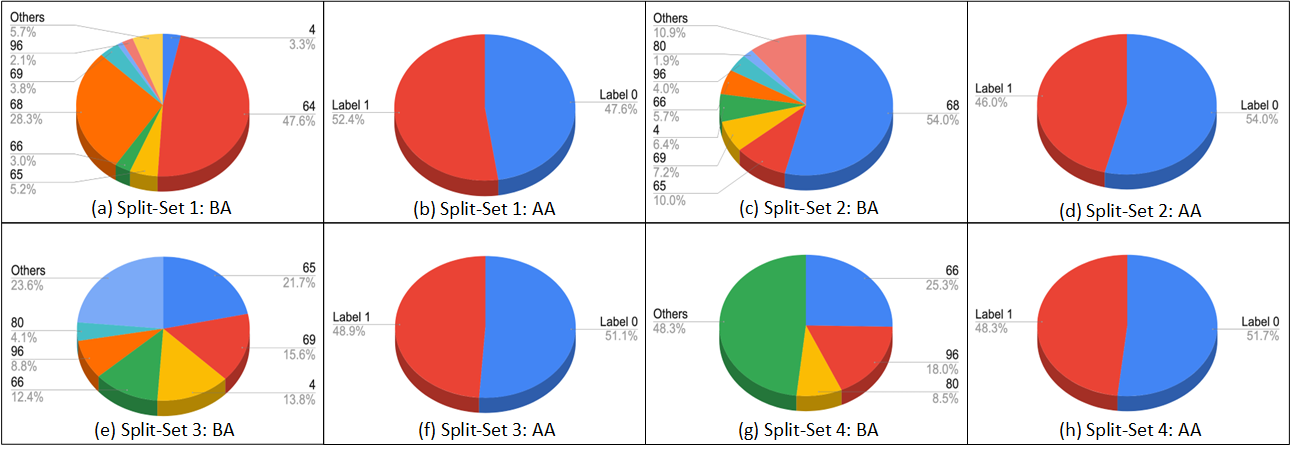}
  \caption{Label distribution of 4 models in the dataset before aggregation (BA) and after aggregation (AA)}
  \label{fig:pie}
\end{figure*}

As we have seen above, this hierarchical data modeling framework allows the ML models (M1, M2, M3, and M4) to be trained on the hierarchical split-sets and predict the 5 aggregated groups--from the top to bottom in the tree--that are resulted in the leaves of the binary tree, as presented in Figure \ref{fig:hmodel}. We can observe these five groups of combined emotions frequently occur in conversational dialogues. Hence, these 5 classes are used in the simulation section to train the RF and ANN models.  

\section{LDEB-DailyDialog}
The \emph{Emo\_Sum} in our derived dataset is the prediction variable. By the LDEB technique explained above, we were able to aggregate the annotated emotion classes into one value which signifies the emotion(s) present in the conversational dialogue. Along with our feature space constructed from the occurrence of all the unique words in the dialogues and aggregating multi-classes to compute Emo\_Sum value using the LDEB technique, we created the LDEB-DailyDialog dataset to train machine learning models. We believe that this dataset will be helpful to train not only language models but also large language models by following the approach systematically on emerging datasets.

\section{Simulations}
We extracted four Split-Sets (or independent subdomains) of the LDEB-DailyDialog dataset using our hierarchical machine learning model, presented in Figure \ref{fig:hmodel}. We named these four subsets--associating them with the four hierarchical models M1, M2, M3, and M4--as Split-Set 1, Split-Set 2, Split-Set 3, and Split-Set 4. The label distribution of these Split-Sets is presented in Figure \ref{fig:hmodel}. The actual data domain (rows, columns) of the feature (sub) spaces of the Split-Sets, where "rows" represents the number of dialogues and "columns" represents the number of features (or tokens), are as follows: 

\begin{itemize}
    \item \textbf{Split-Set 1}: \\(13117, 26373)--as illustrated in Figure \ref{fig:pie}(b).
    \item \textbf{Split-Set 2}: \\(6870, 26373)--as illustrated in Figure \ref{fig:pie}(d).
    \item \textbf{Split-Set 3}: \\(3162, 26373)--as illustrated in Figure \ref{fig:pie}(f).
    \item \textbf{Split-Set 4}: \\(1546, 26373)--as illustrated in Figure \ref{fig:pie}(h).
\end{itemize}

There are 26,373 features that make the dimensionality of the feature space (and the subspaces). After the aggregation of the labels, the Split-Sets formed are as follows: (i) Split-Set 1: Emo\_Sum: 64 (label 0 for Split-Set 1) vs remaining Emo\_Sum values (label 1 for Split-Set 1) (ii) Split-Set 2: Emo\_Sum: 68 (label 0 for Split-Set 2) vs remaining Emo\_Sum values (label 1 for Split-Set 2) (iii) Split-Set 3: Emo\_Sum: 65, 69, 4 (label 0 for Split-Set 3) vs remaining Emo\_Sum values (label 1 for Split-Set 3), and (iv) Split-Set 4: Emo\_Sum: 66, 96, 80 (label 0 for Split-Set 4) vs remaining Emo\_Sum values (label 1 for Split-Set 4). The splitting of the subsets is previously shown in Figure \ref{fig:hmodel}. As shown in Figure \ref{fig:pie}, the datasets are now nearly balanced--after aggregation (AA) compared to before aggregation (BA)--to the extent they alleviate the problem of imbalanced data. The number of dialogues of two major classes are as follows:

\begin{itemize}
    \item \textbf{Split-Set 1}: \\ Label 0 (47.6\%) vs Label 1 (52.4\%).
    \item \textbf{Split-Set 2}: \\Label 0 (54.0\%) vs Label 1 (46.0\%).
    \item \textbf{Split-Set 3}: \\Label 0 (51.1\%) vs Label 1 (48.9\%).
    \item \textbf{Split-Set 4}: \\Label 0 (51.7\%) vs Label 1 (48.3\%).
\end{itemize}

These percentages indicate that our hierarchical data model splits the given LDEB-DailyDialog data into balanced Split-Sets with the constraint that dictates the hierarchically ordered application of the models M1, M2, M3, and M4, as in Figure \ref{fig:hmodel}.

\subsection{Confusion Matrix}

\begin{figure*}[t!]
  \centering
  \includegraphics[width=0.99\textwidth]{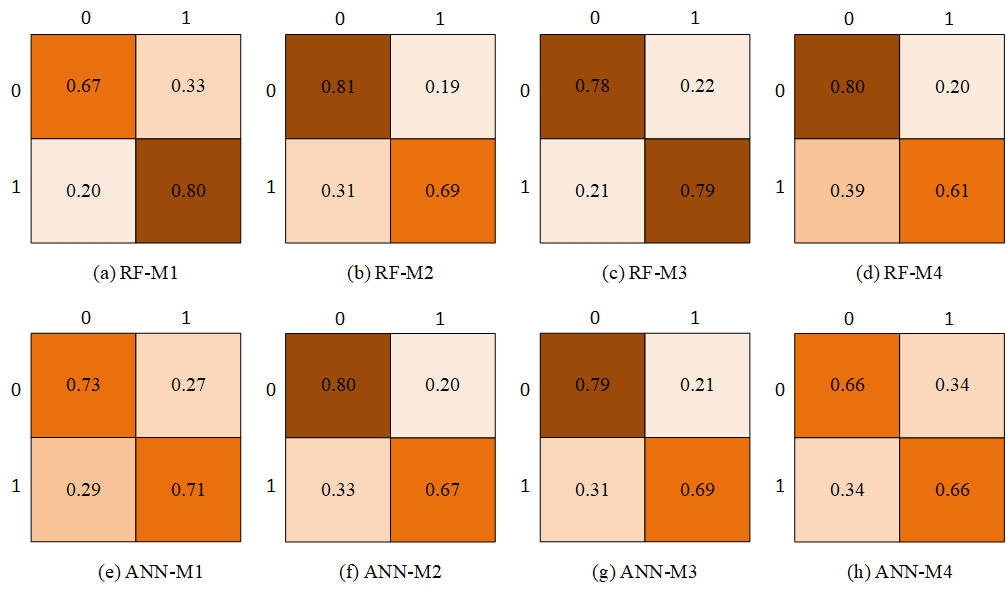}
  \caption{Confusion Matrix of Random Forest (RF) and Artificial Neural Network (ANN) on 4 Models}
  \label{fig:matrix}
\end{figure*}

The confusion matrix in Figure \ref{fig:matrix} shows the predictive performance measures and misclassification errors of RF and ANN when they are used with our proposed hierarchical data modeling. From the data collected from our hierarchical data modeling framework, 80\% of the data has been used as the training data for the ML models and 20\% of the data has been used as test data. The configuration of the ML models are as follows: (a) For RF, we used the default parameters from the sklearn package with \textit{n\_estimators} set to 100 trees. For \textit{criterion}, we selected \textit{gini} - Gini impurity measure - for its simplicity and computational efficiency and since it tends to be less sensitive to class imbalance. For \textit{max\_features}, we opted for \textit{sqrt} which randomly selects a square root number of features for each tree. Finally, we set the \textit{bootstrap} parameter to \textit{True} which offers robustness to overfitting and provides improved generalization so that RF can better capture the underlying patterns and relationships (b) For ANN architecture, we used 891, 828, and 734 that are derived using a simple optimization technique for the neuron in the first three dense layer where we also used \textit{uniform} kernel initializer and \textit{relu} activation function. In the last dense layer, we used 2 neurons. We also compiled the ANN model with stochastic gradient descent optimizer, mean squared error loss, and accuracy as a measure along with the batch size of 20 and epochs of 80.

For example, Figure \ref{fig:matrix}(a) shows that the RF-M1 model predicts 67\% of label 0s as 0s, and 33\% 0s as 1s, while ANN-M1 (Figure \ref{fig:matrix}(e)) model predicts 73\% of label 0s as 0s, and 27\% 0s as 1s. On the other hand, Figure \ref{fig:matrix}(a) shows that the RF-M1 model predicts 80\% of label 1s as 1s, and 20\% 1s as 0s, while ANN-M1 (Figure \ref{fig:matrix}(e)) model predicts 71\% of label 1s as 1s, and 29\% 1s as 0s. We can also observe from Figure \ref{fig:matrix}(b) and Figure \ref{fig:matrix}(f), both RF-M2 and ANN-M2 perform equally well with 81\% and 80\% predictive performance of the positive class at that  node of the hierarchy of the data model. The darkest heatmap of the diagonal cells of the confusion matrices and the higher number of darkest heated cells in the confusion matrices indicate that RF enjoys the hierarchical data model and performs better than the ANN model.

\subsection{Performance Scores}

\begin{table}[t!]
\centering
\setlength{\tabcolsep}{10pt}
\renewcommand{\arraystretch}{1.5}
\begin{tabular}{|c|c|c|c|c|c|}
\hline
 &  & model M1 & model M2 & model M3 & model M4 \\ \hline
RF & A & 0.735 & 0.756 & 0.781 & 0.703 \\ \cline{2-6} 
 & P & 0.719 & 0.753 & 0.743 & 0.744 \\ \cline{2-6} 
 & S & 0.799 & 0.692 & 0.790 & 0.670 \\ \hline
ANN & A & 0.720 & 0.738 & 0.739 & 0.660 \\ \cline{2-6} 
 & P & 0.735 & 0.758 & 0.761 & 0.617 \\ \cline{2-6} 
 & S & 0.708 & 0.671 & 0.690 & 0.657 \\ \hline
\end{tabular}%
\vspace*{5mm}
\caption{It presents test performance scores for the Random Forest classifiers and the Artificial Neural Network classifiers that are associated with the 4 models, where the symbol A represents the accuracy, P represents the precision, and S represents the specificity.}
\label{tab:metrics}
\end{table}

The results presented in Table \ref{tab:metrics} show the comparison of the performance scores of accuracy, precision, and sensitivity between the RF and ANN classifiers after performing them on the 4 Split-Sets, hierarchically, \citep{suthaharan2016supervised}. We ran the experiments on the 4 models several times to obtain the results of the best submission. From our performance analysis, we found that RF performs better than ANN for all the models based on the test accuracy score of 0.735, 0.756, 0.781, and 0.703 for the models M1, M2, M3, and M4 respectively. On the other hand, ANN achieved a slightly lower test accuracy score of 0.720, 0.738, 0.739, and 0.660 for the 4 models, respectively. While the accuracy scores are acceptable to explain the model's performance in our simulations--since we nearly balanced the dataset hierarchically--the precision score is a preferred measure with the interpretation of the sensitivity score, \citep{suthaharan2016supervised}.

For ANN-M1, the precision score of 0.735 ($\sim$0.74) suggests that the TP is nearly 3 times (i.e., 2.85 times) the FP. At the same time, the sensitivity score of 0.708 ($\sim$0.71) suggests that the TP is nearly two-and-a-half times (2.45 times) the FN. It indicates--as far as the positive class is concerned--the ANN-M1 can be considered precise and reasonably less sensitive to the negative class. On the other hand, for RF-M1, the precision score of 0.719 ($\sim$0.72) and the sensitivity score of 0.799 ($\sim$0.80) suggest that this model is less precise and strongly less sensitive to the negative class than the ANN-M1 model. For ANN-M2, the precision score of 0.758 ($\sim$0.76) suggests that the TP is more than 3 times (i.e., 3.2 times) the FP. At the same time, the sensitivity score of 0671 ($\sim$0.67) suggests that the TP is more than 2 times the FN. It indicates--as far as the positive class is concerned--the ANN-M2 can be considered precise and less sensitive to the negative class. On the other hand, for RF-M2, the precision score of 0.753 ($\sim$0.75) and the sensitivity score of 0.692 ($\sim$0.69) suggest that this model is less precise and less sensitive to the effect of the negative class than the ANN-M2 model. For ANN-M3, the precision score of 0.761 ($\sim$0.76) also suggests that the TP is more than 3 times (i.e., 3.2 times) the FP. At the same time, the sensitivity score of 0.69 suggests that the TP is more than twice (2.22 times) the FN. It indicates--as far as the positive class is concerned--the ANN-M3 can be considered precise and less sensitive to the negative class. On the other hand, for RF-M3, the precision score of 0.74 and the sensitivity score of 0.79 suggest that this model is less precise  and significantly less sensitive to the negative class than the ANN-M3 model. In contrast, ANN-M4's performance is significantly low compared to the RF-M4 model, as we can see from the much lower precision and sensitivity scores of ANN-M4. This is understandable since the ANN models require sufficient training samples to boost their performance, but we can see the presence of data paucity for model M4 in the hierarchical model in Figure \ref{fig:hmodel}. 

\section{Discussion}
The simulations also showed us very high training accuracy for the ANN classifier. For example, the ANN classifier delivered accuracy scores of 0.9438 (40 epochs) for model M1, 0.9758 (60 epochs) for model M2, 0.9478 (60 epochs) for model M3, and 0.9515 (100 epochs) for model M4. However, the test accuracy dropped significantly compared to the training accuracy. It indicates there is a significant distribution drift between the training and test sets. In other words, it indicates that the DailyDialog data set has the limitation of supplying sufficient information (or feature vectors) to help the proposed approach perform superior. 

\section{Conclusion}
We were able to define an emotion entanglement problem in conversational dialogues and develop techniques to extract a feature space from the conversational dialogues in a meaningful way to disentangle the twisted association between feature descriptors and the emotion types and predict the emotions in conversations.
We were also able to create a derivative of the DailyDialog dataset--that we called the LDEB-DailyDialog dataset--by implementing label-digitization and emotion-binarization techniques. We hope that the LDEB-DailyDialog dataset will be a great resource for developing improved machine-learning techniques for emotion detection in conversational dialogues. We were also able to detect the missing combinations of emotions in the DailyDialog dataset and this limitation will be addressed in our future research while incorporating additional information to our feature vectors that include positional encoding and word sense disambiguation. We will also add more conversations to alleviate some of the shortcomings of the DailyDialog dataset. We also learned that the techniques proposed in the paper can be scaled to accommodate more emotion classes and the hierarchical framework can be revised to balance conversational dialogue datasets.

\section*{Limitations}
The presence of data paucity in the DailyDialog dataset is one of the limitations. The poor performance of ANN, compared to RF, may be considered as the measure of data paucity. In general, neural network models require a large number of data points (i.e., dialogues for this study) in order to learn sufficiently so that they can provide good predictive performance. Another limitation is that the DailyDialog dataset does not have all the possible combinations of emotions that are defined by the  LDEB approach; hence, the proposed research is limited to the available emotions. However, its flexible nature can allow us to add more emotions to the training as they become available. The limitation also includes the restriction on the combination of the emotions that are imposed by the proposed hierarchical data modeling based on the information available in the DailyDialog dataset.

\bibliographystyle{unsrtnat}
\bibliography{references}  %%% Uncomment this line and comment out the ``thebibliography'' section below to use the external .bib file (using bibtex) .





% \end{thebibliography}

\end{document}